# Smoothness-based Edge Detection using Low-SNR Camera for Robot Navigation

**Vu Hoang Minh** · **Tajwar Abrar Aleef** · **Usama Pervaiz** · **Yeman Brhane Hagos** · **Saed Khawaldeh**

**Abstract** In the emerging advancement in the branch of autonomous robotics, the ability of a robot to efficiently localize and construct maps of its surrounding is crucial. This paper deals with utilizing thermal-infrared cameras, as opposed to conventional cameras as the primary sensor to capture images of the robot's surroundings. For localization, the images need to be further processed before feeding them to a navigational system. The main motivation of this paper was to develop an edge detection methodology capable of utilizing the low-SNR poor output from such a thermal camera and effectively detect smooth edges of the surrounding environment. The enhanced edge detector proposed in this paper takes the raw image from the thermal sensor, denoises the images, applies Canny edge detection followed by CSS method. The edges are ranked to remove any noise and only edges of the highest rank are kept. Then, the broken edges are linked by computing edge metrics and a smooth edge of the surrounding is displayed in a binary image. Several comparisons are also made in the paper between the proposed technique and the existing techniques.

## 1 Introduction

The importance of computer science has grown rapidly in recent decades. Machine vision is one such field of computer science which aims at extracting and analyzing useful information from images and videos for industrial and medical applications, for example, medical imaging [17], signature identification [26], optical character recognition [27], handwriting recognition [30],

Vu Hoang Minh
E-mail: hoangminh.vu@smartdatics.com

object recognition [14], pattern recognition [7], Simultaneous Localization and Mapping (SLAM) [6] and so on.

Feature detection is a fundamental process in image processing which is one part of machine vision. There are four common types of image features including edges, interest points, interest regions and ridges. Feature detection is a technique with the purpose of detecting interest points inferring the contents of an image. To be more specific, these interest points are distinctive in the shape of isolated points, and they show significant differences in all directions in the neighborhood. A large number of feature detection approaches have been proposed in the literature. For example, an intensity-based method based on the auto-correlation of the signal was proposed by Moravec in 1980 [20]; a corner-score-based technique regarding direction directly was developed by Harris and Stephens in 1988 [9]; and a real-time machine learning approach was developed by Rosten and Drummond in 2006 [23]. Feature points of an image are used for image registration [11], object detection [28] and classification [8], tracking [24], and motion estimation [31].

The development of this research work is concerned with the robustness and tractability issues of thermal-infrared image and video. The principal contributions of this research work focus on the enhancement of existing approaches; the proposal of a number of novel techniques in low Signal-Noise Ratio (SNR) image processing; and the evaluation of existing and proposed methods.

The generation of effective feature tracking results using visible-optical cameras is still an active field of research, though quite mature. However, the use of a thermal-infrared camera in similar issue is relatively new and challenging. Recently the usage of thermal-infrared cameras in research has increased due to their ability to detect normally unseen objects by obtaining radiation information in conditions where even a high-end camera fails. For example, low light environment, foggy environment, etc. From this identified problem, this research work has been motivated by two questions:

– How can a conventional camera be replaced by a thermal-infrared camera in normal conditions?
– How to handle the disadvantages of a thermal-infrared camera, namely low-SNR output and poor resolution?

By investigating the capability of thermal-infrared video in a feature detection and tracking system, this research work is contributing to the performance of SLAM and robot navigation systems in unknown environments.

In this research work, a few popular approaches in the field of image processing are studied. In particular, this research work is focused on low-SNR thermal-infrared imagery. One of the core objectives of the research work is to investigate the fundamental differences between visual and thermal-infrared modalities with the purpose of proposing an effective methodology for feature detection and tracking in thermal-infrared video. It can be seen that most current techniques aim at the visual modality, so they do not perform consistently in thermal-infrared images. Therefore, the objectives of this research work are:

– Explore the effects of noise reduction filters in low-SNR videos and propose one technique to improve edge detection.
– Propose an effective, specialized feature detection algorithm for low-SNR thermal-infrared images.
– Propose an edge-strength ranking process in a standard framework.

The rest of this article is organized as follows. Section 2 gives a background of Thermal-infrared modality. Next, Section 3 relates to Canny detector and Global and Local Curvature Properties detector. In Section 4, a novel edge detector using thermal-infrared camera is introduced. After that, results of common edge detectors and proposed detector are shown and compared to check the effectiveness of this research work in Section 5. Finally, Section 6 summarizes and concludes the article.

## 2 Background

### 2.1 Thermal-infrared Radiation

#### 2.1.1 Categorization

Objects at a temperature greater than absolute zero (0 degrees Kelvin) produce thermal radiation. The temperature affects directly the wavelengths of this form of Electro-Magnetic (EM) radiation. This leads to a scheme classified by wavelength [18]:

– Near Infrared (NIR): from 0.7 to 1.0 $\mu m$
– Short-Wave Infrared (SWIF): 1.0 to 3.0 $\mu m$
– Mid-Wave Infrared (MWIF): 3.0 to 5.0 $\mu m$
– Long-Wave Infrared (LWIF): 8.0 to 12.0 $\mu m$
– Very Long-Wave Infrared (VLWIF): 12.0 to 30.0 $\mu m$

In the context of digital output, LWIF is normally referred to as thermal infrared; while NIR is used interchangeably with the term visibility to human eye.

Visible and infrared subsections of the EM spectrum is shown in Figure 1[1].

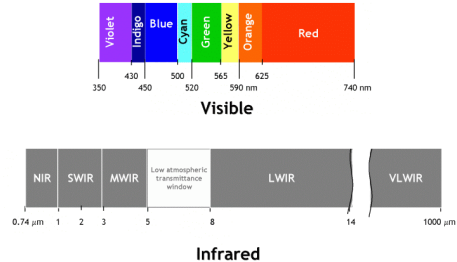

**Fig. 1** Visible and infrared subsections of the EM spectrum

The wavelength of the radiation influences the absorptivity, reflectivity, and emissivity of all objects. In fact, the property of thermal radiation is expressed by Kirchoff's Law [3]. In his research, Beiser et al indicated that under normal circumstances, most of the materials do not reflect much LWIF. Therefore, only emitted radiation originating from the material itself is detected. This phenomenon principally explains why thermal-infrared images are less affected by changes in the environment, such as to lighting, shadows and out-of-view motion.

---

[1] Figure 1 is a modification of one contributed by user Magnus Manske under the CC BY-SA 3.0 license at http://en.wikipedia.org/wiki/File:Infrared_spectrum.gif

### 2.1.2 Electromagnetic Radiation

All objects eject radiation as a function of their materials and temperature. Objects that ideally absorbs all radiation is referred as blackbodies. Planck's Law discusses clearly the EM radiation of a blackbody object [15].

$$B_\lambda(\lambda, T) = \frac{2hc^2}{\lambda^5} \frac{1}{e^{\frac{hc}{\lambda k_B T}} - 1} \qquad (1)$$

Here $B$ is the spectral radiance of the blackbody surface, $T$ is absolute temperature, $\lambda$ is the wavelength, $k_B$ is the Boltzmann constant, $h$ is the Planck constant and $c$ is the speed of light.

Wien's Displacement Law (see Figure 2[2]) also states that:

$$\lambda_{\max} = \frac{b}{T} \qquad (2)$$

Here $T$ is the absolute temperature in Kelvin, and $b$ is Wien's displacement constant.

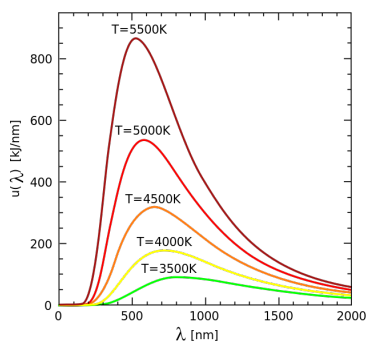

**Fig. 2** Wien's Law of radiation

This law relates to a number of everyday experiences.

- Objects with a surface temperature similar to that of the sun - in dark red color - have peak radiant emittance
- Objects at orange-red temperature, for example, a piece of metal heated by a torch (4500 to 5000 degree Kelvin), predominate radiant emittance and are invisible.

- Objects at temperature 4000 degrees Kelvin, such as a light bulb, human body, and many everyday machines emit a considerable amount of radiation energy and are detectable by thermal-infrared cameras.
- At 1500 degrees Kelvin, the campfire is for warming purpose but not visibility.

## 2.2 Thermal-infrared Cameras

To a certain extent, the operation of thermal-infrared cameras is similar to visible light cameras. However, in term of capturing thermal-infrared radiation, thermal-infrared cameras are complex in design and expensive, but produce very low resolutions. In general, there are two types of thermal-infrared cameras depending on detectors they use: quantum or thermal [13].

### 2.2.1 Optris PI450

A Optris PI450 thermal-infrared camera is used in this research work. Its specifications is shown in Figure 3 [3].

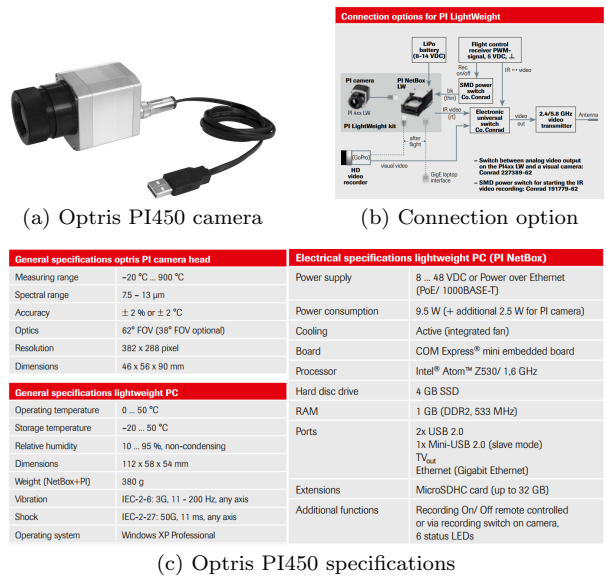

(a) Optris PI450 camera    (b) Connection option

(c) Optris PI450 specifications

**Fig. 3** Thermal-infrared camera and specifications

### 2.2.2 Noise

Image noise is the effect of error in data acquisition of thermal-infrared camera. It may arise from the sensor

---

or from the bolometer pixels. In thermal-infrared imagery context, there is a phenomenon called Fixed Pattern Noise (FPN) which is small but unavoidable. FPN is affected by two factors: Dark Signal Non-Uniformity (DSNU) and Photo Response Non-Uniformity (PRNU). It is more challenging as non-uniformities vary over time.

In practical applications, for example, SLAM, images with FPN are unusable. With the purpose of recovering the original thermal-infrared signal, initial calibration or on-going real-time processing or a combination of both methods is employed. This research work only focuses on real-time image processing. In specific, a natural image denoising is used as a process of removing noise from a corrupted image.

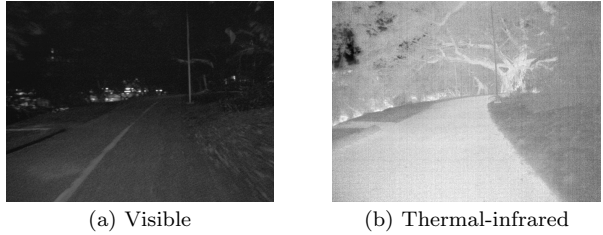

(a) Visible      (b) Thermal-infrared

**Fig. 4** Comparison of visible and thermal-infrared images at same time and condition

### 2.3 Thermal-infrared Imagery

In principle, thermal-infrared images can be viewed and processed in much the same way as regular greyscale images captured from a conventional visible spectrum camera. However, there are several important differences between images in these two modalities. Considering these differences can ensure that the strengths of thermal-infrared images are not squandered, and that the weaknesses are effectively managed.

#### 2.3.1 Variation in SNR

Basically, thermal-infrared cameras produce low SNR images [12], which is why thermal-infrared output seems to be noisy. When the temperature of an object increases, it produces more light and the wavelength of this light becomes shorter, therefore FPN swamps the useful information in an image. Consequently, in these conditions where there are only passive objects, the SNR of thermal-infrared image is low.

Figure 5 demonstrates the variation in SNR of thermal-infrared imagery that occurs between different viewpoints within the same environment.

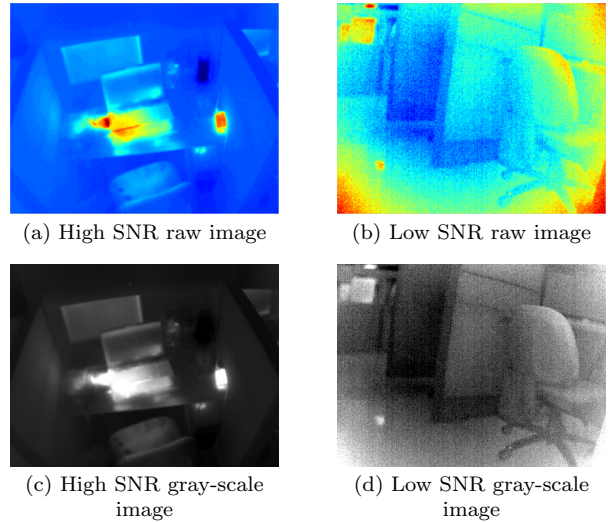

(a) High SNR raw image      (b) Low SNR raw image

(c) High SNR gray-scale image      (d) Low SNR gray-scale image

**Fig. 5** Example of high and low SNR images

#### 2.3.2 Degrees per graylevel

It is worth to get a clear concept about DPG. The raw 14-bit images from the Optris have a Degrees Per Graylevel (DPG) of 0.01. This means that each change in graylevel in the output image represents 0.01 degrees Celsius. Therefore, a difference in 100 graylevels will represent 1 degree Celsius. Hence, for example, if the input image has graylevels ranging from 2217 to 3342, that will mean it has a range of temperatures of:

$$\frac{3342 - 2217}{100} = 11.25°C \qquad (3)$$

This would be a fairly high SNR image. A low SNR image with graylevels ranging from 2560 to 2692 would represent a range of temperatures of:

$$\frac{2692 - 2560}{100} = 1.32°C \qquad (4)$$

Within one raw 16-bit sequence, the temperature range of each image may change, but the DPG stays the same: fixed at 0.01. Depending on the method of converting from 14-bit to 8-bit images, the DPG may be modified. In fact, many conversion methods will result in a different DPG for every output image in the sequence, which is not desirable. Effectively, a lower DPG means a higher magnitude of noise.

Hence, an approach that will maintain the same DPG in the 8-bit output image, even when the SNR is very different, is recommended to use. For example, one approach is to find the median intensity of the 14-bit image, and then scale the raw graylevels by a fixed factor, such as 0.5, and threshold to keep all values within the range of $0 - 255$. The pseudo code of scaling algorithm is shown in Algorithm 1.

---

**Algorithm 1** Scale DPG

---

1: **procedure** **scaledpg**(image)
2:    ***input***: raw image $R$
3:    find median value $M$ of $R$
4:    **for** every pixel has intensity $R(x, y)$ **do**
5:       $A = [R(x, y) - I] * 0.5 + 127$
6:       $B = \min(A, 255)$
7:       $G(x, y) = \max(B, 0)$
8:    **end for**
9:    ***output***: scaled image $G$
10: **end procedure**

---

For low SNR images, these may appear to a human viewer to have very low contrast. But for image processing and edge detection algorithms, they are able to be more sensitive. The important thing is to use the same formula for converting all types of images to 8-bit. Therefore, a script to generate a directory of 8-bit images from raw 16-bit images, and using the default scaling will ensure the same DPG for all output images is expected.

## 3 Related works

### 3.1 Canny Detector

Canny edge detector was developed by Canny in 1986 in a paper named "A Computational Approach To Edge Detection" [4]. This technique is considered one of the most successful edge detection methods and is widely used in the engineering applications. Canny defined three criteria of good edge detection. They are good detection-which is the ability that an algorithm marks actual edges, good localization- which shows how near the marked edge is to the real edge, and minimal response-which expresses noise should not construct false edges.

Canny edge detector was enhanced by Bao et al by improving localization by using scale multiplication in 2005 [2]. In 2009, an improved Canny edge detection was proposed by Wang [29]. In specific, Wang propounded using self-adaptive filter instead of Gaussian one to refine the original approach.

The Canny algorithm is demonstrated in six major steps.

*Step 1*

Since noisy and unprocessed raw image is the input, Canny edge detector uses Gaussian blur to filter out noise. A convolution mask, which is much smaller than the raw image, is slid over the image. The greater the size of Gaussian kernel is, the less sensitive the detector to noise. Hence, the pick of window size is very important. An example of $5 \times 5$ Gaussian kernel is demonstrated as following.

$$K = \frac{1}{159} \begin{bmatrix} 2 & 4 & 5 & 4 & 2 \\ 4 & 9 & 12 & 9 & 4 \\ 5 & 12 & 15 & 12 & 5 \\ 4 & 9 & 12 & 9 & 4 \\ 2 & 4 & 5 & 4 & 2 \end{bmatrix} \qquad (5)$$

*Step 2*

After smoothing the raw image, gradient and direction of gradient of each pixel are computed. This step is similar to Sobel method.

$$G_x = \begin{bmatrix} -1 & 0 & 1 \\ -2 & 0 & 2 \\ -1 & 0 & 1 \end{bmatrix} * I \quad \text{and} \quad G_y = \begin{bmatrix} 1 & 2 & 1 \\ 0 & 0 & 0 \\ -1 & -2 & -1 \end{bmatrix} * I \quad (6)$$

*Step 3*

Gradient is computed as.

$$\nabla I(x, y) = G(x, y) = \sqrt{G_x^2(x, y) + G_y^2(x, y)} \qquad (7)$$

*Step 4*

Gradient's direction is followed.

$$\Theta(x, y) = \arctan\left( \frac{G_y(x, y)}{G_x(x, y)} \right) \qquad (8)$$

The direction is rounded to one of four feasible angles, namely $0, 45, 90$ and $135$ degrees, and is illustrated in Figure 6.

*Step 5*

After gradient's direction is found, non-maximum suppression is applied to acquire thin lines in binary image. Non-maximum suppression considers every pixel given the image gradients.

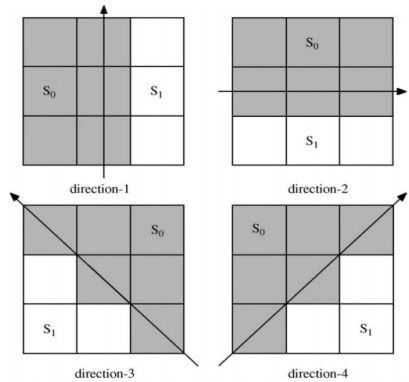

**Fig. 6** Four direction of two sets of pixels in $3 \times 3$ mask [21]

*Step 6*

Final step of Canny detector is hysteresis, which involves high and low thresholds. In specific, a pixel is considered belonging to an edge if its pixel gradient is higher than the high threshold. Otherwise, a pixel is considered not belonging to an edge if its pixel gradient is smaller than low threshold. If a pixel gradient is between the two thresholds, it is accepted belonging to an edge if and only if it is connected to a pixel having gradient above the high threshold.

*Pseudo-code*

The pseudo-code of Canny edge detector is shown in Algorithm 2.

### 3.2 Global and Local Curvature Properties Detector

Global and Local Curvature Properties Detector (GLCPD) is published in Optical Engineering journal in May 2008. This work was proposed by He and Yung in an article named "Corner detector based on global and local curvature properties" [10]. GLC performs well in terms of feature correspondence and subjective manner. The algorithm was developed based on an early technique, Curvature Scale Space (CSS), proposed by Mokhtarian and Suomela in 1998 [19].

Compared to the original CSS method, GLC evidence of the abilities to detect true corners and eliminate false ones. In addition, GLC outperforms the inspired technique by improving localization in a small neighborhood. According to He and Yung, the main contributions of their approach are to detect feature points by computing global and local curvature properties to distinguish round corners from obtuse corners and to parameterize the approach.

---

**Algorithm 2** Canny edge detector

1: **procedure cannyedge**(image)
2:    *input*: image
3:    apply Gassian blur to remove noise
4:    **for** every pixel in the image **do**
5:      compute gradient
6:      compute direction of gradient
7:    **end for**
8:    relate gradient direction    ▷ 0,45,90 or 135 degrees
9:    **for** every pixel in the image **do**
10:     suppress non-maximum
11:    **end for**
12:    **for** every pixel in the image **do**
13:      **if** gradient is greater than high threshold **then**
14:        this pixel is included in binary image
15:      **else if** gradient is less than low threshold **then**
16:        this pixel is NOT included in binary image
17:      **else**
18:        **if** connected to a pixel having gradient greater than high threshold **then**
19:          this pixel is included in binary image
20:        **else**
21:          this pixel is NOT included in binary image
22:        **end if**
23:      **end if**
24:    **end for**
25:    *output*: binary image
26: **end procedure**

---

GLC shows a promising performance, in spite of its dependence on an edge detector, namely Canny method, in preprocessing step. Intuitively speaking, GLC is a powerful technique with regard to capabilities of distinguishing true corners and optimizing localization errors.

He and Yung developed GLC with the purpose of utilizing global and local curvature properties. With this philosophy, a new procedure is proposed based on [19].

*Step 1*

Apply Canny detector to obtain a binary edge map. A comprehensive process of Canny detector is shown in Section 3.1.

*Step 2*

According to CSS method proposed by Mokhtarian and Suomela, contours are extracted by Algorithm 3.

*Step 3*

Next, He and Yung proposed to compute the curvature value of each pixel of contours to find the true corners.

---
**Algorithm 3** CSS algorithm

---
**procedure extractcontour**(binary image)
    ***input***: binary image
    **for** every point not be marked **do**
        store index
        set current point to be this point
        **while** not connected to end-point **do**
            fill the gap
            change current point to filled point
        **end while**
        **if** connected to end-point **then**
            extract contour
        **end if**
        **if** end-point connected to an extracted edge **then**
            set as T-junction
        **else**
            set as end-point
        **end if**
    **end for**
    ***output***: contours
**end procedure**

---

Curvature value of pixel $i$ in contour $j$ is given by.

$$K_i^j = \frac{\Delta x_i^j \Delta^2 y_i^j - \Delta^2 x_i^j \Delta y_i^j}{[(\Delta x_i^j)^2 + (\Delta y_i^j)^2]^{1.5}} \quad \text{for } i = 1, 2, ..., N \quad (9)$$

*Step 4*

According to He and Yung, Region of Support (ROS) is defined by the portion of a contour bounded by two nearest curvature minimas. Consequently, an adaptive threshold in an ROS is calculated as follows.

$$T(u) = R \times \overline{K} = R \times \frac{1}{L_1 + L_2 + 1} \sum_{i=u-L_2}^{u+L_1} |K_i| \quad (10)$$

where $L_1 + L_2$ is the size of ROS, centered at $u$, $R$ denotes the minimum ratio of major axis to minor axis of an ellipse (default value of $R$ is 1.5), and $\overline{K}$ is the mean curvature of the ROS.

Round corners are removed if their curvature values are smaller than the adaptive thresholds in ROS. Otherwise, corners are defined as true corners.

*Step 5*

Next step is to remove false corners. He and Yung mention that a well-defined corner should have a relatively sharp angle. It means obtuse corners, which are greater than $\theta_{obtuse} = 162°$, should be eliminated. To alleviate this problem, [10] determine the angle of an arbitrary point as.

$$\angle C = \begin{cases} |\gamma_1 - \gamma_2| & \text{if } |\gamma_1 - \gamma_2| < \pi \\ 2\pi - |\gamma_1 - \gamma_2| & \text{otherwise} \end{cases} \quad (11)$$

where $\gamma_1$ and $\gamma_2$ are the tangents from this point to two end-points of ROS.

*Step 6*

Finally, a contour

$$A^j = \{P_1^j, P_2^j, ..., P_N^j\} \quad (12)$$

is determined to be closed (*loop*) or open (*line*).

$$A^j \begin{cases} closed & \text{if } \overline{|P_1^j, P_N^j|} < T \\ open & \text{if } |P_1^j, P_N^j| > T \end{cases} \quad (13)$$

*Pseudo-code*

To summarize, the pseudo-code of GLC algorithm is illustrated in Algorithm 4.

---
**Algorithm 4** GLCP detector

---
1: **procedure curvaturecorner**(image)
2:   ***input***: an image
3:   extract binary image by Algorithm 2
4:   extract contours from binary edge map by Algorithm 3
5:   **for** every pixel in contours **do**
6:     compute curvature value by Equation 9    ▷ $K_i^j$
7:   **end for**
8:   **for** each pixel in contours **do**
9:     compute adaptive threshold in ROS by Equation 10    ▷ $T(u)$
10:     **if** $K_i^j < T(u)$ **then**
11:       remove round corner
12:     **end if**
13:   **end for**
14:   **for** each corner **do**
15:     compute angle of corner    ▷ $\angle C$
16:     **if** $\angle C > \theta_{obtuse}$ **then**
17:       remove false corner
18:     **end if**
19:   **end for**
20:   ***output***: corner points
21: **end procedure**

---

## 4 Proposed method

A novel edge detection method is proposed to improve on existing Canny edge detector. This algorithm can enhance the robustness and quality of output binary image by removing most of the fake edges generated by noise. This technique is shown to be effective in low-SNR video.

### 4.1 Outline

The proposed detector has been implemented in Matlab using a combination of open source libraries provided by Matlab society. It has been contrived as a process which takes in the raw image as input and outputs a binary image with a collection of indexed tracked edges.

The proposed technique is inspired by Canny edge detector [4] and Global and Local Curvature Properties Detector [10]. The performance is promising since most of the fake edges are filtered out.

### 4.2 Algorithm

There are seven major steps in the proposed smoothness-based detector.

#### *Step 1*

The first step is to denoise by Gaussian kernel or well-engineered noise reduction filters, such as Non-local means (NL-means) [1] and Sparse 3D Transform-domain Collaborative Filter (BM3D) [5].

The output after Step 1 is demonstrated in Figure 7 and Figure 8. For high-SNR image (Figure 7), there is no big differences between raw and denoised images. However, in low SNR scenario (Figure 8), BM3D illustrates the best output and uniformly performs better than the others, This is because its output contains structured objects which allow a productive grouping and collaborative filtering.

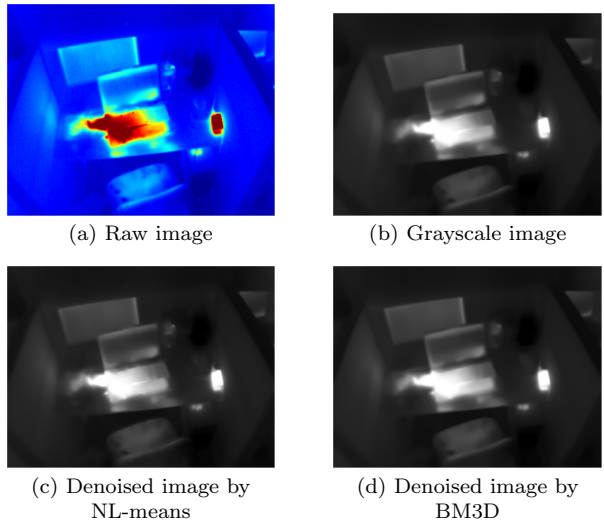

(a) Raw image          (b) Grayscale image

(c) Denoised image by          (d) Denoised image by
NL-means                         BM3D

**Fig. 7** Outputs of proposed edge detection after Step 1 for high SNR image

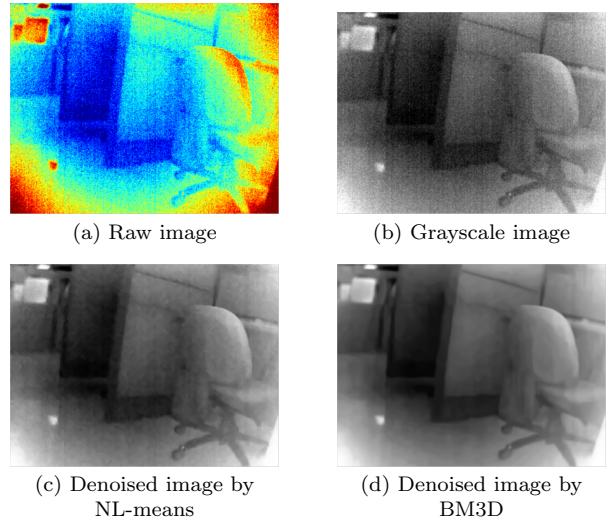

(a) Raw image          (b) Grayscale image

(c) Denoised image by          (d) Denoised image by
NL-means                         BM3D

**Fig. 8** Outputs of proposed edge detection after Step 1 for low SNR image

#### *Step 2*

The second step is to apply Canny algorithm (see Section 3.1) to detect edges. Figure 9 and 10 show binary images after being denoised. BM3D once again shows its effectiveness in terms of preservation of details and reduction of noise.

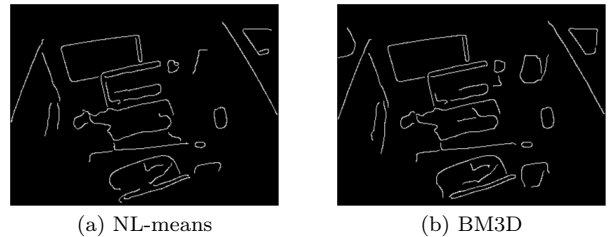

(a) NL-means          (b) BM3D

**Fig. 9** Outputs of proposed edge detection after Step 2 for high SNR image. The inputs of (a) and (b) are taken from Step 1 using Nl-means and BM3D, repectively

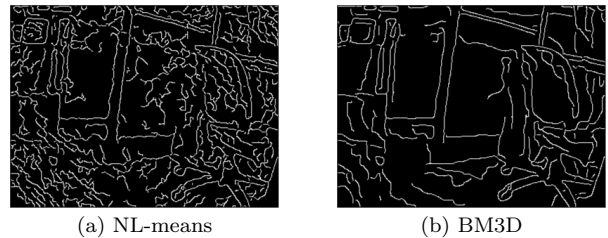

(a) NL-means          (b) BM3D

**Fig. 10** Outputs of proposed edge detection after Step 2 for low SNR image. The inputs of (a) and (b) are taken from Step 1 using Nl-means and BM3D, repectively

*Step 3*

Next, binary edge map found from Step 2 is the input of Step 3 [10] in order to extract contours as in the CSS method [22]. After that, some steps are essential to follow in order to find true corners and remove false ones. Details are broached in Section 3.2. According to Rattarangsi and Chin, six steps in the CSS method are demonstrated below [10].

1. Detect binary map.
2. Extract edges.
3. Compute curvature for each edge.
4. Remove round corners by comparing to an adaptive threshold.
5. Remove false corners.
6. Set end points as corners.

   True corners in different type of images are shown in Figure 11 and 12.

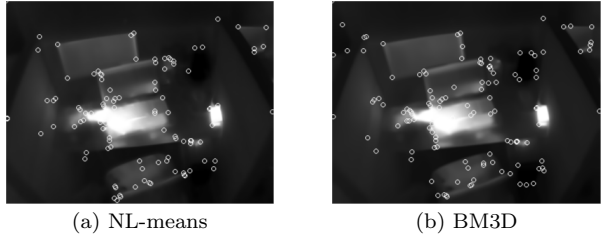

| (a) NL-means | (b) BM3D |

**Fig. 11** Outputs of proposed edge detection after Step 3 for high SNR image. The inputs of (a) and (b) are taken from Step 2 using Nl-means and BM3D, repectively

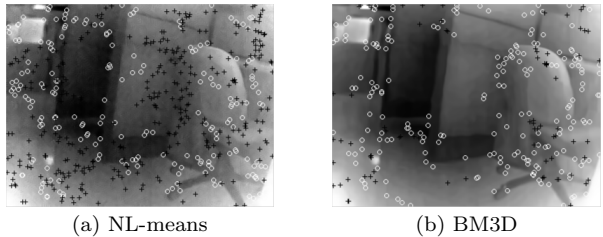

| (a) NL-means | (b) BM3D |

**Fig. 12** Outputs of proposed edge detection after Step 3 for low SNR image. The inputs of (a) and (b) are taken from Step 2 using Nl-means and BM3D, repectively

*Step 4*

This step is to link edges provided from Step 3 and update corners accordingly. There are three types of linking.

– Type 1: the ending of an edge to the ending of another edge. This type of linking combines two edges into a longer one.
– Type 2: the ending of an edge to a midpoint of another edge. Here, this type creates a T-junction.
– Type 3: the beginning to the ending of an edge. This type of linking changes the curve mode from *line* to *loop*.

To recall, a contour

$$A^j = \{P_1^j, P_2^j, ..., P_N^j\}$$

is determined to be closed (*loop*) or open (*line*).

$$A^j \begin{cases} closed & \text{if } \overline{|P_1^j, P_N^j|} < T \\ open & \text{if } |P_1^j, P_N^j| > T \end{cases}$$

By default, T is set at 2-3 pixels. The algorithm of Step 4 is shown in Algorithm 5.

---

**Algorithm 5** Link edges
---
1: **procedure** linkedge
2:     **input**: binary image, GAP_LINK
3:     extract contours from binary image
4:     **for** every edge in an image **do**
5:         **for** each point in an edge, A **do**
6:             **if** exist a pixel, B, where distance A-B is smaller than GAP_LINK **then**
7:                 **if** A and B are beginning points of different edges **then**
8:                     Type-1
9:                     update edges
10:                    update corners
11:                 **else if** A is mid point and B is beginning point **then**
12:                     Type-2
13:                     update edges
14:                     update corners
15:                 **else if** A and B are beginning and ending points of an edge **then**
16:                     Type-3
17:                     update curve-mode to *loop*
18:                     update edges
19:                     update corners
20:                 **end if**
21:             **end if**
22:         **end for**
23:     **end for**
24:     **output**: binary image with contours
25: **end procedure**

---

*Step 5*

Next step is to compute edges' metrics. This step is described as follows.

Firstly, for the given contours and corners in Step 4, which corner belongs to which contour should be done.

At the end of this sub-step, a table showing a list of edges with corners lying on them is expected. Based on the selection of ending point is corner [10], therefore, there are greater or equal to two corners on each edge.

For each edge, based on its corners, its metric is computed. An edge metric depends on four aspects.

- Length of this edge. The longer the edge is, the less possible it is generated by noise.
- Smoothness of this edge. The detected edge generated by the object is more likely to be smoother than the one generated by noise.
- Number of corners lying on this edge.
- Curve mode. *Line* or *loop*. This factor represents how much a loop is preferable than a line.

Consider an N-pixels edge,

$$A^j = \{P_1^j, P_2^j, ..., P_N^j\}$$

with $M$ corners in order.

$$C^j = \{C_1^j, C_2^j, ..., C_M^j\}$$

Clearly, $M \leq N$, $P_1^j = C_1^j$ and $P_N^j = C_M^j$.

The strategy of computing edge-score is summarized as follows.

1. From 2 adjacent corners, extract a small curve between them.
2. Fit the extracted curve to a cubic model.

$$f(x) = ax^3 + bx^2 + cx + d \tag{14}$$

3. Compute summation of squares of residuals.

$$RSS_j = \sum_{i=1}^{N}(y_i^j - f^j(x_i))^2 \tag{15}$$

where $y_i$ is real index (of extracted edge), and $f(x_i)$ is estimated index (of fitted model).

4. Compute summation of all summations (RSS) of divided parts.

$$RSS = \sum_{j=1}^{M-1} RSS_j \tag{16}$$

5. Compute average residual (AR).

$$AR = \frac{RSS}{N} \tag{17}$$

6. Edge-Score (ES) is computed as.

$$ES = \frac{AR \times W \times L \times M^2}{N^2 \times N' \times \Phi} \tag{18}$$

Here $W$ and $L$ are width and length of image, respectively; $N$ is number of pixels of an edge; $M$ is number of corners lying on this edge; $AR$ is average residual; $N'$ is the summation of pixels of all edges connected to this edge and this edge; and $\Phi$ is loop-line-ratio.

*Step 6*

Next, edges are ranked based on their metrics. The smaller the metric, the better the edge.

*Step 7*

Finally, $ET$ strongest edges are kept, the weaker are removed. $ET$ represents a threshold of the number of edges which is desirable to be kept.

*Pseudo-code*

The flowchart and pseudo-code of Proposed Smoothness-based Detector is shown in Figure 13 and Algorithm 6, respectively.

## 5 Results

As edges are fundamental elements in machine vision, it is crucial to evaluate edge detectors. Here, two low SNR thermal-infrared images are used for evaluation. Figure 14 shows a workspace including a number of objects, such as a monitor, laptop, charger, mobile phone and so on; while Figure 15 illustrates the same space but at a different viewpoint.

It can be seen from Figure 14 and 15 that classical detectors, such as Prewitt, Roberts and Sobel, are very sensitive to noise and inaccurate, especially Roberts' in Figure 14. In particular, a lot of short edges or even dots are detected that look like impulse noise. The more noise appears or the lower the magnitude of the edges is, the worse final result is [25] [16]. Hence, these edges should be removed. However, classical operators are very simple since they only use two steps of convolutions. Furthermore, the basics of detected edges with orientations are sufficient in some circumstances.

Laplacian of Gaussian detector demonstrates a better performance compared to classical operators although

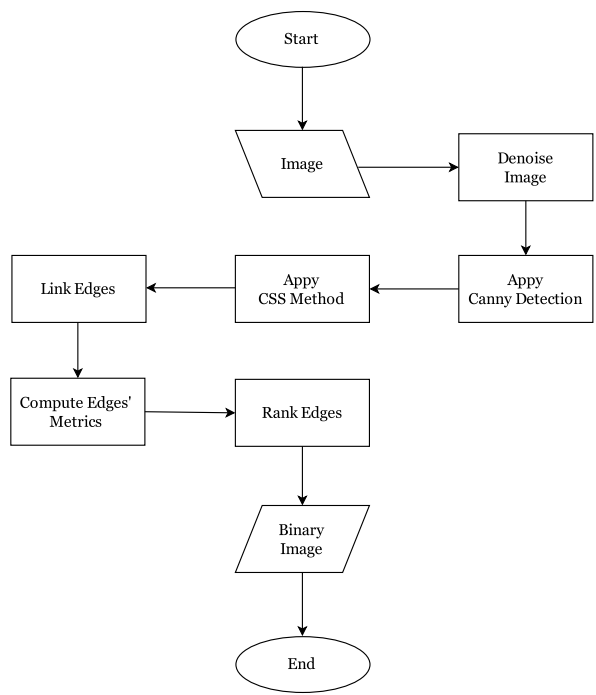

**Fig. 13** Flowchart of Enhanced Edge Detector

**Algorithm 6** Edge Detection: Proposed Smoothness-based Detector

1: **procedure smoothnessbasededge**(image)
2:    ***input***: image, $ET$ - edge threshold
3:    filter noise
4:    apply Canny edge detection to find binary image
5:    apply Global and Local Curvature Properties Detector to find corners
6:    link edges
7:    find which corner belong to which edge
8:    **for** every edge in the provided binary image **do**
9:       extract this edge into $M - 1$ parts
10:       **for** every part **do**
11:          fit this part to a cubic polynomial         ▷
   $ax^3 + bx^2 + cx + d$
12:          find $a$, $b$, $c$ and $d$
13:          estimate $y$ index given $x$ index
14:          compute summation of squares of residuals
15:       **end for**
16:       compute summation of all summations of squares of residuals
17:       compute average residual
18:       compute edge-score
19:    **end for**
20:    rank edges
21:    **if** $ET \geq N$ (number of edges) **then**
22:       keep all edges
23:    **else**
24:       keep $ET$ strongest edges
25:    **end if**
26:    ***output***: binary image
27: **end procedure**

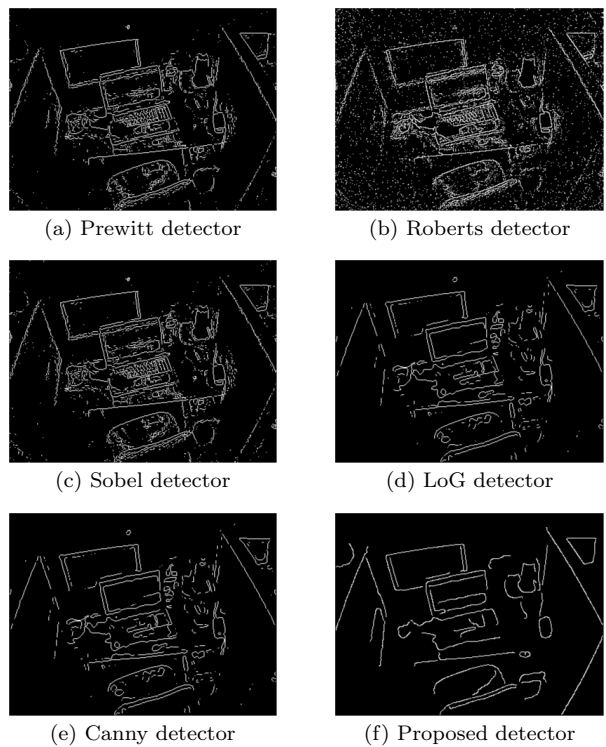

(a) Prewitt detector      (b) Roberts detector

(c) Sobel detector      (d) LoG detector

(e) Canny detector      (f) Proposed detector

**Fig. 14** Detection results on denoised high SNR image shown in Figure 7

it applies standard convolution methods. The advantages of Laplacian of Gaussian detector are the capability to find the correct location of edges and the use of wider Gaussian kernel, $9 \times 9$; hence, the outcome binary image is better than classical operators. But, Laplacian of Gaussian detector fails to find gradients of edges and curves having a variety of intensities.

Canny's algorithm is still considered one of the most successful edge detection methods and is widely used in the engineering applications. However, the Canny detector can be considered as computationally complex and time-consuming since it has six steps (see Section 3.1).

The proposed algorithm is an enhanced Canny detector. It can be seen from Figure 14 and Figure 15 that the proposed approach removes most of the edges made by noise. Moreover, it preserves the details even at two regions having very low-intensity differences. Consequently, it outperforms all the other methods in terms of connected edges and structured object detection. But, similar to the Canny method, it also has a complex computation.

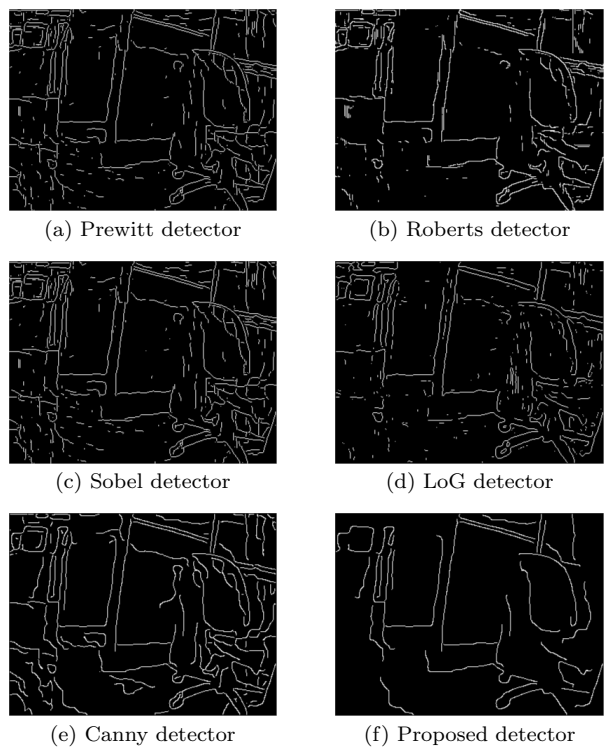

(a) Prewitt detector      (b) Roberts detector

(c) Sobel detector      (d) LoG detector

(e) Canny detector      (f) Proposed detector

**Fig. 15** Detection results on denoised low SNR image shown in Figure 8

## 6 Conclusion

For robots designed for operating in conditions with almost no access to any light source, the utilization of thermal-infrared cameras can greatly assist in the navigation of the robot. In this paper, a novel algorithm for edge detection is proposed for efficiently detecting edges of images taken from thermal-infrared cameras. The main goal of this work was to introduce an efficient edge detection method which concentrates on filtering the excess edges and output the vital ones.

Some of the key points of the proposed methodology are as follows:

- Explore the effects of noise reduction filters in low-SNR videos.
- Propose an effective, specialized feature detection algorithm for low-SNR thermal-infrared images.
- Propose an edge-strength ranking process in a standard framework.

Future works could include using conventional cameras for the evaluation of this research work in normal conditions. Furthermore, an enhanced feature detector can be developed based on this methodology by proposing feature ranking technique.